\documentclass{article}
\usepackage{ijcai25}

\usepackage{times}
\usepackage{soul}
\usepackage{url}
\usepackage[hidelinks]{hyperref}
\usepackage[utf8]{inputenc}
\usepackage[small]{caption}
\usepackage{graphicx}
\usepackage{amsmath}
\usepackage{amsthm}
\usepackage{booktabs}
\usepackage{algorithm}
\usepackage{algorithmic}
\usepackage[switch]{lineno}

\usepackage{array}
\usepackage{multirow}
\usepackage{amssymb}

\title{Diffusion Models for Computational Neuroimaging: A Survey}

\author{
Haokai Zhao$^1$
\and
Haowei Lou$^1$\and
Lina Yao$^{1,2}$\and
Wei Peng$^{3}$\and
Ehsan Adeli$^{3}$\and
Kilian M Pohl$^{3}$\and
\\Yu Zhang$^{4,5}$\\
\affiliations
$^1$Computer Science Building (K17), Engineering Rd, UNSW Sydney, Kensington NSW 2052, Australia\\
$^2$CSIRO's Data61, Level 5/13 Garden St, Eveleigh NSW 2015, Australia\\
$^3$Department of Psychiatry and Behavioral Sciences, Stanford University, Stanford, CA 94305, USA\\
$^4$Department of Bioengineering, Lehigh University, Bethlehem, PA 18015, USA\\
$^5$Department of Electrical and Computer Engineering, Lehigh University, Bethlehem, PA 18015, USA\\
\emails
haokai.zhao@student.unsw.edu.au, 
haowei.lou@student.unsw.edu.au,
lina.yao@data61.csiro.au,
wepeng@stanford.edu,
eadeli@stanford.edu,
kilian.pohl@stanford.edu, 
yuzi20@lehigh.edu
}

\begin{document}

\maketitle

\begin{abstract}
Computational neuroimaging involves analyzing brain images or signals to provide mechanistic insights and predictive tools for human cognition and behavior. While diffusion models have shown stability and high-quality generation in natural images, there is increasing interest in adapting them to analyze brain data for various neurological tasks such as data enhancement, disease diagnosis and brain decoding. This survey provides an overview of recent efforts to integrate diffusion models into computational neuroimaging. 
We begin by introducing the common neuroimaging data modalities, follow with the diffusion formulations and conditioning mechanisms. Then we discuss how the variations of the denoising starting point, condition input and generation target of diffusion models are developed and enhance specific neuroimaging tasks. For a comprehensive overview of the ongoing research, we provide a publicly available repository at https://github.com/JoeZhao527/dm4neuro.
\end{abstract}

\section{Introduction}
With the development of neuroimaging technologies and computational resources in recent decades, there is a growing research community that uses computational methods that include statistical, machine learning, and deep learning \cite{avbervsek2022deep} to improve the understanding of the human brain and build tools for the prediction of clinical variables based on neuroimaging data. However, challenges persist due to the high costs associated with data collection, the inherent high-dimensionality of neural data, and the complex, noisy, and inhomogeneous nature of the data itself. Existing methods, while powerful, still face limitations in addressing these issues, leaving room for further refinement and innovation.

Diffusion models, which exhibit stability, realistic sample generation, and a robust ability to model underlying data distributions \cite{ho2020denoising,song2020score}, are receiving growing attention for their potential to solve the challenges in computational neuroimaging. As a class of generative models, they can be used to generate realistic brain images or signals, helping to overcome issues such as small sample sizes and noisy data, thus improving downstream analyses \cite{pinaya2022brain}. Moreover, with variations in their formulation, conditioning mechanisms, and generation targets, diffusion models are showing promise in enhancing performance in a range of brain analysis tasks. For instance, in data enhancement tasks like reconstruction and super-resolution, diffusion models have potentials to provide missing semantic details from under-sampled signals or low-resolution brain images with their generation abilities compared to the deterministic approaches, while ensuring the anatomical and functional reality with their stability. In neural disease diagnosis, diffusion models can provide counterfactual generations, and their differences can be used for brain tumor segmentation or pathological analysis. In the field of brain decoding, brain signals can be incorporated with the diffusion models pre-trained on natural images or speeches, yield high quality decoding. Exploration of the aligned feature space of brain signals and external stimulus can provide insights for correspondence of brain activations and the external inputs.

\begin{figure*}
    \centering
    \includegraphics[width=1\linewidth]{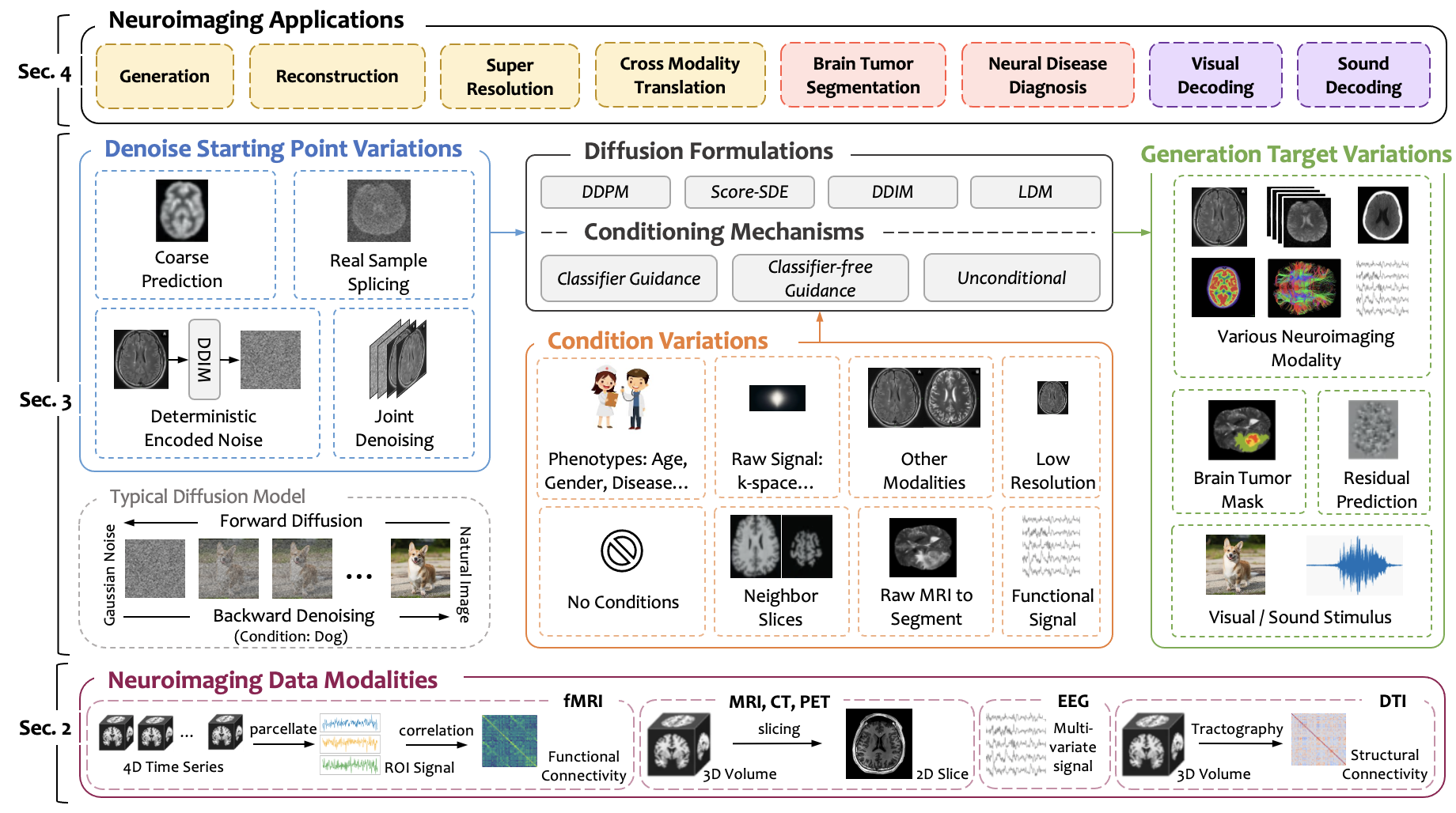}
    \caption{An overview of  diffusion models for neuroimaging. Based on fundamental diffusion formulations and conditioning mechanisms, variations on denoise starting point, input condition and generation target are developed for different neuroimaging data and tasks.}
    \label{fig:overview}
\end{figure*}

Despite the growing interest in applying diffusion models to neuroimaging, a comprehensive review of their specialized applications in this domain remains lacking. Our survey addresses this gap by providing a focused examination of how tailored designs and formulations of diffusion models enhance various neurological tasks.  From the application perspective, we categorized the applications of diffusion models in neuroimaging into eight major fields, including generation, reconstruction, super-resolution, cross-modality translation, brain tumor segmentation, neural disorder diagnosis, visual decoding, and speech decoding. Moreover, our survey covers extended neuroimaging modalities, including Magnetic Resonance Imaging (MRI), Functional Magnetic Resonance Imaging (fMRI), Computed Tomography (CT), Diffusion Tensor Imaging (DTI), Electroencephalogram (EEG) and Positron Emission Tomography (PET). An overview of our survey is illustrated in Figure \ref{fig:overview}.

This survey is structured as follows. To introduce the essential background, we first cover the commonly used data modalities in neuroimaging in Section \ref{sec:data}, and introduce the foundational concepts of diffusion models and their popular variations of formulations in Section \ref{sec:theory}. Then, we provide a detailed discussion on the designs that focus on developing diffusion models specifically for solving neuroimaging problems in Section \ref{sec:application}. Finally, the open challenges and future directions are discussed in Section \ref{sec:future}.

\section{Neuroimaging Data Modalities} \label{sec:data}
Neuroimaging data collected with different techniques has distinct characteristics, formats, and processing methods, offering insights into the brain's structure and function. Below, we provide an overview of several prominent neuroimaging modalities, including the format of data they produce and common processing steps.

\textbf{Magnetic Resonance Imaging (MRI)} produces high-resolution 3D structural images of the brain, with each voxel representing tissue types like gray matter, white matter, and cerebrospinal fluid. In some cases, 2D slices are extracted from the 3D volume for easier analysis. MRI data are often used for anatomical analysis, detecting structural abnormalities, and providing a baseline for other modalities.

\textbf{Functional Magnetic Resonance Imaging (fMRI)} measures brain activity through blood oxygenation level-dependent (BOLD) signals. The data is 4D, consisting of a sequence of 3D volumes over time, which allows for studying dynamic brain activity in response to stimuli. For downstream analysis, fMRI data are often processed as time series of Region-of-Interest (ROI) signals, where the brain is divided into functional regions. These ROI signals can then be used to construct functional connectivity by computing correlations between ROIs over time.

\textbf{Computed Tomography (CT)} provides 3D brain images using X-rays, capturing tissue density and structural information. While it has lower spatial resolution than MRI, CT is fast and useful for detecting acute conditions such as hemorrhages, tumors, or strokes.

\textbf{Diffusion Tensor Imaging (DTI)} is a specialized form of MRI. It maps the diffusion of water molecules in the brain. It generates 3D tensor fields that describe diffusion directionality and strength, enabling the reconstruction of white matter tracts (structural connectivity) and the assessment of white matter integrity.

\textbf{Electroencephalogram (EEG)} captures electrical brain activity via scalp electrodes. It offers high temporal resolution (millisecond precision) but low spatial resolution. EEG data is typically represented as time series for each electrode, reflecting voltage fluctuations associated with brain activity.

\textbf{Positron Emission Tomography (PET)} is a nuclear imaging technique that measures functional brain activity by tracking the uptake of radioactive tracers. These tracers bind to specific molecules or receptors, allowing measurement of processes such as glucose metabolism or neurotransmitter release. PET scans are 3D images showing tracer distribution in the brain.

\section{Diffusion Models} \label{sec:theory}

\subsection{Diffusion Formulations} \label{sec:formulation}
Diffusion models are a class of generative models that learn to gradually transform a simple distribution (e.g., Gaussian noise) into complex data through a series of diffusion steps. These models have been adapted into various formulations to improve their performance and applicability across different domains. In this section, we discuss the foundational formulations and several popular variations of diffusion models.

\textbf{Denoising Diffusion Probabilistic Models (DDPM)} \cite{ho2020denoising} represents a foundational framework for diffusion models. They utilize a two-step Markovian process: forward diffusion and reverse denoising. In the forward process, Gaussian noise is incrementally added to the data over multiple time steps, progressively transforming the data distribution into a standard Gaussian distribution. This transformation is governed by a sequence of predefined variances that control the noise levels at each step.

The reverse denoising process aims to recover the original data by iteratively removing the added noise. The neural network is trained to predict and subtract the noise at each step. The training objective minimizes the difference between the predicted and actual noise, effectively teaching the model to denoise the data.

\textbf{Score-Based Generative Models (Score-SDE)}, also known as Noise Conditional Score Networks \cite{song2019generative}, take a distinct approach by estimating the score function of the data distribution, i.e., the gradient of the log probability density with respect to the data. By learning this score function across various noise levels, these models gain the ability to denoise data corrupted by different levels of noise.

Different from DDPMs, this approach can be formulated using Stochastic Differential Equations (SDEs) \cite{song2020score}, providing a continuous-time framework for diffusion models. The forward SDE describes how data is diffused continuously over time, while the reverse-time SDE, which incorporates the learned score function, enables data generation.

\textbf{Denoising Diffusion Implicit Models (DDIM)} \cite{song2021denoising} reformulates the diffusion process in DDPMs to significantly accelerate the generative process. Theoretically, \cite{song2021denoising} begins by defining a non-Markovian forward process and a corresponding trainable generative process, which replace the diffusion process in DDPM. It demonstrates that both the forward and backward processes can be deterministic, and that generation in this framework is approximately the same regardless of the number of forward steps. This formulation not only enables faster generation with fewer diffusion steps but also retains high-level features regardless of the generation trajectory. In terms of implementation, DDIM modifies only the generation steps and removes the added noise in each step compared to DDPM.

\textbf{Latent Diffusion Models (LDMs)} \cite{rombach2022high} enhance both computational efficiency and modeling accuracy by operating in a lower-dimensional latent space that captures the most salient features of the data. Using an autoencoder, the high-dimensional data is compressed into this latent representation, focusing the diffusion process on the most informative aspects. This significantly reduces the computational costs of training and sampling.

\subsection{Conditioning Mechanisms} \label{sec:condition}
Diffusion models can be either conditional \cite{ho2022classifier,dhariwal2021diffusion} or unconditional \cite{ho2020denoising}. While unconditional diffusion models can serve as general priors for data distributions, conditioning on health status, biological traits, or cross-modal references significantly enhances the adaptability of diffusion models to specialized tasks. There are two primary methods for conditioning diffusion processes: classifier-free guidance \cite{ho2022classifier} and classifier guidance \cite{dhariwal2021diffusion}, which we refer to ”conditional training” and ”conditional inference” in this survey.

\textbf{Conditional training} incorporates conditioning information directly into the diffusion model during training. A prominent method is classifier-free guidance, where conditioning information is randomly omitted with a certain probability during training. This strategy allows the model to learn both conditional and unconditional generation, providing flexibility during inference. The advantage of classifier-free guidance is that it eliminates the need for a separate classifier at inference time. The model is trained to handle conditioning directly, making it simpler to adjust the strength of guidance by interpolating between conditional and unconditional predictions. 

\textbf{Conditional inference} introduces conditioning during the sampling phase of a trained unconditional diffusion model. In classifier guidance, a separate classifier is trained to estimate the probability of a target condition given a noisy sample. During sampling, the diffusion process is guided by the gradients of the classifier's output with respect to the sample, adjusting the denoising trajectory toward the desired condition. In the field of neuroimaging reconstruction, super-resolution, and cross-modality translation, common practices involve using raw brain signals, low-resolution brain images, and cross-modality references to guide the inference process instead of the classifiers.

\subsection{Task Related Variations} \label{sec:task_variations}

In addition to the diffusion formulations and conditioning mechanisms, significant adaptations are made to diffusion models to cater to different neuroimaging tasks. These adaptations primarily revolve around three key aspects: the denoising starting point, conditioning input, and generation target. While the conditioning input and generation target are often tightly linked, we will discuss them together, as they define the role of the diffusion model in a particular task. The following section outlines how each of these variations plays a crucial role in task-specific adaptation, with further detailed applications discussed in Section \ref{sec:application}.

\textbf{Denoising Starting Point:} Traditional diffusion models begin the generation process with random Gaussian noise \cite{ho2020denoising}. However, such random noise can lead to variability in the generation results and produce unrealistic brain structures. To address this, \cite{zhao2025diffusion} proposes using a noisy real sample as the starting point for functional connectivity generation, ensuring biological relevance in the generated samples. Furthermore, with the deterministic nature of denoising in DDIM, the endpoint of the diffusion process, which can be seen as a latent space, is also leveraged as a starting point for cross-modality translation tasks. For instance, the diffusion endpoint of one modality can serve as the denoising starting point for generating another modality, thereby enhancing task-specific coherence \cite{luo2024target}.

\textbf{Condition Input and Generation Target:} The condition input and generation target in diffusion models are often inherently linked. These inputs and targets are typically defined based on the specific task at hand. For example, in super-resolution tasks, low-resolution data is used as input (conditioning), while high-resolution data serves as the target output \cite{li2024rethinking,wang2023inversesr}. Similarly, in brain decoding, brain signals are used as input, and the output might be natural images or speech, which are the generation targets \cite{takagi2023high,liu2024reverse}. In some tasks, such as brain tumor segmentation, the model's output can be further processed to derive task-specific results. For instance, \cite{pinaya2022fast} utilizes the difference between healthy and abnormal samples generated by the model to produce segmentation outcomes.

\begin{table*}[ht]
    \centering
    \addtolength{\leftskip} {-2cm}
    \addtolength{\rightskip}{-2cm}
    \tiny
    \renewcommand{\arraystretch}{1.4}
    \begin{tabular}{
        >{\centering\arraybackslash}m{2.6cm}
        >{\centering\arraybackslash}m{1.4cm}
        >{\centering\arraybackslash}m{1.2cm}
        >{\centering\arraybackslash}m{0.5cm}
        >{\centering\arraybackslash}m{0.5cm}
        >{\centering\arraybackslash}m{1.2cm}
        >{\centering\arraybackslash}m{2.5cm}
        >{\centering\arraybackslash}m{2.5cm}
        >{\centering\arraybackslash}m{2.5cm}
    }
        \hline
        \textbf{Model} & \textbf{Task} & \textbf{Form.} & \textbf{Train Cond.} & \textbf{Inf. Cond.} & \textbf{Modality} & \textbf{Denoise Starting Point} & \textbf{Condition Inputs} & \textbf{Generation Target} \\ \hline

        \cite{pinaya2022brain} & \multirow{4}{*}{Generation} & LDM & \checkmark &  & MRI & Gaussian Noise & Age/Sex covariates & 3D MRI \\ 
        \cite{peng2023generating} & & DDPM & \checkmark & & MRI & Gaussian Noise & Neighbour slices & 3D MRI \\ 
        \cite{aristimunha2023synthetic} & & LDM & & & EEG & Gaussian Noise & - & EEG \\
        \cite{yuan2024diffusion} & & DDPM & & & fMRI & Gaussian Noise & - & fMRI signal \\ \hline

        \cite{song2021solving} & \multirow{3}{*}{Reconstruction} & Score-SDE & & \checkmark & MRI, CT & Gaussian Noise & Sinogram/k-space & 2D MRI \\ 
        \cite{tu2023wkgm} & & Score-SDE & & \checkmark & MRI & Gaussian Noise & k-space & k-space \\ 
        \cite{han2023contrastive} & & DDPM & \checkmark & & PET & Corase Prediction & Neighbour slices, spectrum & 2D PET \\ \hline

        \cite{wang2023inversesr} & \multirow{4}{*}{Super Resolution} & LDM, DDIM & & \checkmark & MRI & Gaussian Noise & LR MRI & HR MRI \\ 
        \cite{mao2023disc} & & LDM & \checkmark & & MRI & Gaussian Noise & LR MRI & HR MRI \\ 
        \cite{li2024rethinking} & & LDM & \checkmark & & MRI & Gaussian Noise & LR MRI & HR MRI \\ 
        \cite{zhou2024spatio} & & LDM & \checkmark & & EEG & Gaussian Noise & LR EEG & HR EEG \\ \hline

        \cite{kim2024adaptive} & \multirow{3}{*}{Cross Modality} & LDM & \checkmark & & MRI & Gaussian Noise & Source modality & Target modalities \\ 
        \cite{jiang2023cola} & & LDM & \checkmark & & MRI & Gaussian Noise & Source modalities & Target modality \\
        \cite{luo2024target} & & DDIM & & \checkmark & MRI, CT & Noise encoded from DDIM & Domain classifier & Target modality \\ 
        \cite{meng2024multi} & & DDPM & & & MRI & Noise + source modality & - & Target modalities \\ \hline

        \cite{zhou2023generative} & \multirow{4}{*}{Disorder Diagnosis} & LDM & \checkmark & & EEG & Gaussian Noise & Alzheimer’s labels & EEG \\ 
        \cite{zhao2025diffusion} & & LDM & \checkmark & & fMRI & Noisy Real Sample & ASD's labels & Functional connectivity \\ 
        \cite{zong2024new} & & LDM & \checkmark & & DTI & Gaussian Noise & Reference and moving DTI & Deformation field \\ 
        \cite{zong2024brainnetdiff} & & LDM & \checkmark & & DTI, fMRI & Gaussian Noise & fMRI signal & Structural connectivity \\ \hline

        \cite{wu2024medsegdiff} & \multirow{3}{*}{Segmentation} & LDM & \checkmark & & MRI & Gaussian Noise & MRI & Segmentation mask \\ 
        \cite{li2024corrdiff} & & LDM & \checkmark & & MRI & Gaussian Noise & MRI & Error map \\ 
        \cite{pinaya2022fast} & & LDM & & & MRI & Gaussian Noise & MRI & MRI \\ \hline

        \cite{takagi2023high} & \multirow{3}{*}{Visual Decode} & LDM & \checkmark & & fMRI & Gaussian Noise & fMRI signal & Natural image \\ 
        \cite{chen2023seeing} & & LDM & \checkmark & & fMRI & Gaussian Noise & fMRI signal & Natural image \\ 
        \cite{luo2024brain} & & LDM & \checkmark & & fMRI & Gaussian Noise & fMRI signal & Natural image \\ \hline

        \cite{liu2024reverse} & Speech Decode & LDM & \checkmark & & fMRI & Gaussian Noise & fMRI signal & Speech spectrogram \\ 
        \hline 
    \end{tabular}
    \caption{A summary of diffusion model applications in neuroimaging, organized by task type (generation, reconstruction, super-resolution, cross-modality translation, neural disorder diagnosis, brain tumor segmentation, visual decoding and speech decoding). Key aspects include diffusion formulations (Form.), conditional training/inference (Train cond./Inf. Cond), data modalities, denoising starting point, condition inputs, and generation targets. LR and HR in super-resolusion refer to low-resolution and high-resolution, respectively.}
    \label{tab:model_summary}
\end{table*}

\section{Applications} \label{sec:application}

In this section, we discuss key applications of diffusion models in neuroimaging, including brain data generation, reconstruction, super-resolution, cross-modality translation, brain tumor segmentation, neural disorder diagnosis, visual and sound decoding (Table. \ref{tab:model_summary}). For each application, we first present a brief problem definition, then introduce several representative works and discuss their key designs of task-specific diffusion models.

\subsection{Data Generation}
Data generation aims to produce synthetic samples that resemble real data while offering enhanced diversity. Such samples are useful for data augmentation and improving downstream analyses. Research in this area typically focuses on generation fidelity and diversity, evaluated using metrics like Fréchet Inception Distance (FID) and Structural Similarity Index Measure (SSIM).

One of the key challenges in generating brain images is their high dimensionality, which introduces heavy computation loads. \cite{peng2023generating} provides a memory-efficient solution. It trains a 2D DDPM to generate MRI subvolumes conditioned on another arbitrary subset of slices from the same MRI. At the inference time, the 3D brain MRI is generated by repeatedly applying the 2D slices generation. Alternatively,\cite{pinaya2022brain} generates 3D T1-weighted MRI images by using LDM to compress the raw MRI to a much lower dimension latent space. The LDM was conditioned on age, sex, and brain structure volumes to ensure correctness with specific covariates.

Beyond MRI, diffusion models have been adapted to brain signals including EEG and ROI signals from fMRI. In \cite{aristimunha2023synthetic}, an unconditional LDM is trained to generate sleeping stage EEG signals. It applies an extra spectral loss to ensure the realistic neural
oscillation and could. For fMRI ROI signal, \cite{yuan2024diffusion} develops a DDPM that directly performs generation on signal space. The model explicitly disentangles trend, periodic, and residual components using Fourier-based synthetic layers and transformer architectures to ensure the realistic and interpretable generation.

\subsection{Reconstruction}

Reconstruction involves generating high-quality brain images from undersampled data. Traditional reconstruction methods often rely on solving inverse problems, where the reconstruction process is deterministic and the quality of the reconstructed image is proportional to the sampling frequency. Diffusion models bypass these limitations by learning the underlying data distributions.

A representative framework for brain MRI and CT reconstruction is to train a diffusion model to learn the prior on high-quality images, then use undersampled raw signal as a guidance in the generation process. In \cite{song2021solving}, the score-SDEs were applied to solve inverse problems, reconstructing images from sinograms and k-space data. The score-SDE is trained to model the MRI/CT image prior, and the sampling process is done by a proximal optimization step to enforce consistency with observed measurements and leveraging predictor-corrector samplers.

Alternatively, \cite{tu2023wkgm} performed MRI reconstruction follows the same paradigm of score-SDE with predictor-corrector sampler, but the generation target is high-resolution k-space signals instead of the MRI image. Furthermore, it effectively reduces dynamic range disparities and making training more stable, by applying a weighted k-space learning strategy.

Besides MRI, PET benefits from diffusion models for reconstructing high-dose images from low-dose acquisitions. In \cite{han2023contrastive}, a conditional diffusion framework uses a coarse-to-fine strategy: a coarse prediction module first estimates the high-dose PET from low-dose input, and then an iterative reverse diffusion process refines it by predicting noise components. Additional conditions (neighboring axial slices and spectral features) and a contrastive loss further ensure consistency by aligning noise predictions from the same subject while distinguishing those from different subjects.

\subsection{Super-Resolution}
Super-resolution aims to generate high-resolution brain images from lower-resolution ones, which can aid in detecting smaller brain structures and abnormalities that might be missed at lower resolutions.

InverseSR \cite{wang2023inversesr} represents a conditional inference approach, where unconditional DDIMs trained in the latent space modeled the high-resolution MRI distribution and low-resolution guidance was introduced to iteratively refine high-resolution outputs. In each iteration, the sampled high-resolution MRI was corrupted to a low-resolution version using a deterministic corruption function. The difference between the generated low-resolution MRI and the reference image was used to update the latent representation.

In comparison, low-resolution images can also serve as conditioning inputs during training, enabling super-resolution through attention mechanisms. DisC-Diff \cite{mao2023disc} develops a LDM that utilizes low-resolution MRI and an additional reference to guide the high-resolution MRI generation. Separate encoders were employed to model independent and shared features for each input, and a condition fusion module was applied to fuse the features for conditioning the diffusion model. DiffMSR \cite{li2024rethinking} further improves the low-resolution conditioned LDM, by replacing the convolutional latent space auto-encoding with a specialized transformer that utilizes large window self-attention, successfully reducing the distortion and improving the computation efficiency.

While most super-resolution is performed based on MRI, \cite{zhou2024spatio} developed a spatio-temporal adaptive framework based on diffusion models for the super-resolution of EEG signals. The training process uses paired low- and high-resolution EEG signals, with the low-resolution signal serving as the condition to generate a high-resolution one. For high-resolution generation, the framework incorporates the diffusion transformer \cite{peebles2023scalable} and multiscale 1D convolutions for the denoising process to capture multi-scale temporal features of EEG. For low-resolution conditioning, a transformer-based spatio-temporal module is applied to encode the low-resolution EEG into a high-level representation.

\subsection{Cross-Modality Translation}
Multi-modal imaging is crucial for comprehensive evaluation in neuroimaging analysis, as it combines complementary information from different imaging techniques. Diffusion models have been used to translate between modalities, compensating for the limitations of individual imaging methods.

One popular paradigm is to use the source image or representation as the condition for target modality representation or image generation. \cite{kim2024adaptive} tackles one-to-many MRI translation with a conditional LDM to perform translation between T1-weighted, T1 contrast-enhanced, T2-weighted, and FLAIR MRI. For the auto-encoding stage, it proposes a multiple-switchable block, which dynamically transforms source latents to target-like latents according to style. Then the latent diffusion model is trained for generating target latent representation, with the source latent as the condition. Similarly, a LDM with encoded representations of other MRI contrasts as conditions for the diffusion process was trained in Cola-Diff \cite{jiang2023cola}, for many-to-one MRI translation. The available modalities were encoded separately and fused with structural guidance using an auto-weight adaptation module as the condition for the latent space diffusion of the target modality. The key advantage of Cola-Diff is that it can maximize the chance of leveraging relevant multi-modal information jointly.

While the aforementioned methods either miss out on the joint information with multiple available modalities \cite{kim2024adaptive} or have to train different models for different target modalities \cite{jiang2023cola}, M2DN \cite{meng2024multi} uses an unconditional DDPM for many-to-many MRI translation. The key difference between M2DN and previous methods is that it treats the observed and missing modalities as a unity. The joint synthesis scheme enhances the construction of a common latent space, not only supports arbitrary missing scenarios, but also improves the overall synthesis performance.

Another important draw-back for cross modality translation is most methods rely on paired data for different modalities, which are not available in some scenarios. \cite{luo2024target} performs unpaired brain MRI-CT translation based on Dual diffusion implicit bridges \cite{su2022dual}. It follows the classifier guidance approach, which trains two DDIM for MRI and CT noise-to-image generation, respectively, and one binary classifier to predict the probability gradient of the image input belonging to the target domain. In the sampling process, it first performs the forward noising process to the source image, then generates the target image from the sampled noise under the guidance of the trained classifier.

\subsection{Neural Disorder Diagnosis}
Neural disorders, such as attention deficit hyperactivity disorder (ADHD), autism spectrum disorder (ASD), Alzheimer's disease, and schizophrenia, significantly impact the quality of life. Diffusion models can be used for feature extraction or data augmentation from structural or functional connectivity data to improve neural disorder diagnosis and prognosis.

Existing brain network construction tools have limitations, including dependence on empirical user input, weak consistency in repeated experiments, and time-consuming processes. \cite{zong2024new} introduces a diffusion model that generate a deformation field conditioned on raw DTI and template reference for brain image alignment, brain region registration, and brain network construction. The framework further applies graph contrastive learning based on the constructed brain network to eliminate individual differences unrelated to diseases, achieving a more stable and robust disease diagnosis.

Functional connectivity and structural connectivity that are constructed from fMRI and DTI, provide different information that can improve neural disorder diagnosis performance. While \cite{zong2024new} only focuses on structural information from DTI, \cite{zong2024brainnetdiff} built a LDM that integrates the functional and structural information. It performs structural connectivity generation conditioned on fMRI Region-of-Interests (ROI) signals. The generated brain networks are then used to perform classification for ADHD and its subtypes. 

Another application of diffusion models in neural disorders is data augmentation. In \cite{zhou2023generative}, data augmentation with diffusion models was performed on EEG data for Alzheimer's disease detection. Raw EEG signals were first transformed into frequency spectra using continuous wavelet transformation. The latent diffusion training and generation were performed on the frequency spectra and conditioned on subjects' health status. For fMRI-based neural disorder diagnosis, \cite{zhao2025diffusion} utilizes Diffusion Transformer \cite{peebles2023scalable} to perform functional connectivity generation and augmentation for Autism Spectrum Disorder (ASD) diagnosis. To ensure realistic connectivity and consistent condition generation, it starts the reverse diffusion from a noisy real sample instead of pure Gaussian noise.

\subsection{Tumor Segmentation}
Brain tumors, whether benign or malignant, can be dangerous due to the limited space within the skull, which restricts growth. Tumors near vital brain regions may cause symptoms such as weakness, impaired mobility, vision loss, language difficulties, and memory issues. Diffusion models have been shown to be effective for brain tumor segmentation due to their ability to learn the underlying data distribution.

A growing trend in brain tumor segmentation is using raw MRI images as the condition for diffusion models for tumor mask generation. \cite{wu2024medsegdiff} proposes a diffusion model with a dynamic conditional encoding module to generate segmentation masks. The raw MRI image is used as the condition for the diffusion process, and the denoising and generation are performed on the segmentation mask. Similarly, \cite{li2024corrdiff} develops a LDM conditioned on the raw MRI image, but it is trained to generate the segmentation error map to correct the initial segmentation and reduce the systematic error of the segmentation. In the inference stage, given the MRI image to segment, it first performs a coarse segmentation, then uses the sum of the initial segmentation and the generated error map as the final corrected segmentation. 

Alternatively, diffusion models can generate MRI images instead of segmentation masks for brain tumor segmentation. In \cite{pinaya2022fast}, the latent diffusion model is trained on healthy subjects to learn the distribution of tumor-free MRI images. During inference, the patient's MRI image is encoded, and a forward-backward diffusion process is performed in the latent space. The generated latent representation was decoded into pixel space, and the residual between the patient's real and generated MRI was used for segmentation.

\subsection{Visual Decoding}
Visual decoding refers to the process of reconstructing or synthesizing visual stimuli (e.g., images, scenes) from neural activity. In neuroscience and cognitive science, this task provides a window into how sensory information is represented, processed, and transformed within the brain.

In \cite{takagi2023high}, visual reconstruction using latent diffusion models was achieved by combining an autoencoder, which learns the mapping from fMRI data of the early visual cortex to low-resolution images, with a latent diffusion model conditioned on representations from the higher (ventral) visual cortex. Another approach was presented in \cite{chen2023seeing}, where a conditional latent diffusion model with enhanced sparse masked brain modeling on patched fMRI was used to extract representations from fMRI signals. A double-conditioning mechanism was developed for latent diffusion training on natural images, conditioned on fMRI latent representations.

Another application of diffusion models in visual decoding is identifying brain regions likely to be activated by particular visual objects. BrainDiVE, a model trained to synthesize images predicted to activate a given region in the human visual cortex, was introduced in \cite{luo2024brain}. This model builds on LDM by conditioning the image generation process through fMRI-guided voxel-wise encoders. Specifically, the model perturbs the denoising process based on gradients from a brain encoder, maximizing activation in targeted regions and allowing fine-tuned image synthesis that mirrors brain responses to complex natural scenes.

\subsection{Speech Decoding}
Speech reconstruction focuses on recovering auditory stimuli from brain signals such as fMRI. Although speech decoding handles inherently temporal signals with spectral and linguistic complexity, it shares similar challenges with visual decoding: extracting high-dimensional features from noisy neural data to re-synthesize the original perceptual experience.

A novel coarse-to-fine method that leverages non-invasive fMRI data to progressively decode brain activity into semantic and acoustic features for speech reconstruction was introduced in \cite{liu2024reverse}. The approach utilizes Contrastive Language-Audio Pretraining for coarse semantic decoding, followed by Audio Masked Autoencoder for fine-grained acoustic decoding. A latent diffusion model was employed to generate mel-spectrograms conditioned on neural features, ultimately generating speech.

\section{Future Directions and Open Challenges} \label{sec:future}

While diffusion models have demonstrated potential in addressing data scarcity, complex representations, and high-dimensionality in neuroimaging, several open challenges remain. There are significant opportunities for further exploration and improvement. In this section, we outline potential future directions for diffusion models in neuroimaging.

\subsection{Representation Learning}
Diffusion models were originally developed for data distribution modeling and generation, but recent research has also explored their potential for learning meaningful representations. Unlike traditional generative tasks, representation learning with diffusion models focuses on extracting latent features that improve downstream tasks. For example, \cite{yang2023diffusion} proposed RepFusion, a method that leverages pre-trained diffusion models for non-generative tasks in natural images. Additionally, \cite{chen2024deconstructing} introduced latent Denoising Autoencoders (l-DAE), a method that deconstructs Denoising Diffusion Models (DDMs) to explore the components that influence self-supervised representation learning. However, the application of diffusion models for representation learning in neuroimaging remains underexplored. Given the importance of learned representations for improving downstream tasks and enhancing our understanding of neural functionalities, this is a promising direction for future research.

\subsection{Privacy-Preserving Computation and Federated Learning}
Brain data collection often raises privacy concerns. Federated learning, which trains models on decentralized data without sharing the raw data, is well-suited for addressing these concerns in neuroimaging. It has already shown promise in the Internet of Medical Things \cite{ali2022federated}. Diffusion models, which learn data distributions without memorizing the exact training data, could be integrated with federated learning to build generalizable models while protecting privacy. For instance, \cite{li2024feddiff} developed a federated learning framework using diffusion models for multi-modal and multi-client data fusion, highlighting the potential of combining these approaches. However, diffusion models also pose new security risks, as they may allow adversaries to reconstruct training data from model weights or gradients \cite{gu2024federated}. This presents both opportunities and challenges for privacy-preserving computation in neuroimaging, making it a critical area for future investigation.

\subsection{Integration with Foundation Models}
Foundation models are large, pre-trained models that can be adapted for various downstream tasks. In recent years, foundation models have been developed for images, text, and multi-modal data \cite{rombach2022high}. Neuroimaging research is beginning to explore the use of foundation models for specific tasks, such as brain tumor segmentation \cite{ma2024segment} and mental state decoding \cite{thomas2022self}. The integration of diffusion models with other types of foundation models could enable more generalized and large-scale training, improving performance across a broader range of neuroimaging tasks.

\subsection{Causal Inference}
Causal inference is important for understanding the underlying causes of specific predictions, which is particularly valuable in neuroimaging for analyzing disease mechanisms and brain region functions. Structural causal models represent causal relationships and can improve the interpretability, reliability, and consistency of generative models \cite{zhou2023opportunity}. Recent studies have successfully combined structural causal models with diffusion models to improve interpretability in image generation tasks \cite{augustin2022diffusion}. Applying this integration to neuroscience problems could lead to deeper insights into brain function and pathology, making causal inference a valuable future direction.

\section{Conclusion}
In conclusion, we present a detailed survey for the application of diffusion models for neuroimaging, providing a taxonomy from the application perspective and highlighting how the diffusion model is induced into the family of neuroimaging. We also provide an in-depth analysis of the current models, discuss their potential applications, and highlight the benefits of incorporating diffusion models in neuroimaging. Finally, we discuss research challenges and directions and aim to provide beneficial insights for future researchers in this field.

\newpage
\bibliographystyle{named}
\bibliography{ijcai25}

\end{document}